%% file: main.tex
\documentclass[runningheads]{llncs}
\usepackage{graphicx}

\usepackage{tikz}
\usepackage{comment}
\usepackage{amsmath,amssymb} 
\usepackage{color}

\usepackage[accsupp]{axessibility}  

\usepackage{graphicx}
\usepackage{amsmath}
\usepackage{amssymb}
\usepackage{booktabs}
\usepackage[ruled,linesnumbered]{algorithm2e}
\usepackage{caption}
\usepackage{subcaption}
\usepackage{multirow}

\usepackage{listings}
\usepackage{color}

\definecolor{dkgreen}{rgb}{0,0.6,0}
\definecolor{gray}{rgb}{0.5,0.5,0.5}
\definecolor{mauve}{rgb}{0.58,0,0.82}

\usepackage[pagebackref,breaklinks,colorlinks]{hyperref}

\usepackage[capitalize]{cleveref}
\crefname{section}{Sec.}{Secs.}
\Crefname{section}{Section}{Sections}
\Crefname{table}{Table}{Tables}
\crefname{table}{Tab.}{Tabs.}

\begin{document}

\pagestyle{headings}
\mainmatter
\def\ECCVSubNumber{7856}  

\title{PTQ4ViT: Post-Training Quantization for Vision Transformers with Twin Uniform Quantization} 

\titlerunning{PTQ4ViT}
%
\author{Zhihang Yuan \inst{1,3} $ ^\star$ \and
Chenhao Xue \inst{1} \thanks{First author and Second Author contribute equally to this paper.} \and
Yiqi Chen \inst{1} \and
Qiang Wu \inst{3} \and \\
Guangyu Sun \inst{2} \thanks{Corresponding author}
}
\authorrunning{Z. Yuan et al.}
%
\institute{School of Computer Science, Peking University\\
\email{\{yuanzhihang, xch927027\}@pku.edu.cn}\and
School of Integrated Circuits, Peking University\\
\email{gsun@pku.edu.cn} \and
Houmo AI}
\maketitle

\input{0_abstract}

\input{1_introduction}

\input{2_background_relatedwork}
\input{3_method}

\input{4_experiment}
\input{6_conclusion}
\input{7_acknowledgement}
\input{9_appendix}

\bibliographystyle{splncs04}
\bibliography{egbib}
\end{document}

%% file: 0_abstract.tex
\begin{abstract}
Quantization is one of the most effective methods to compress neural networks, which has achieved great success on convolutional neural networks (CNNs).
Recently, vision transformers have demonstrated great potential in computer vision.
However, previous post-training quantization methods performed not well on vision transformer, resulting in more than 1\% accuracy drop even in 8-bit quantization.
Therefore, we analyze the problems of quantization on vision transformers.
We observe the distributions of activation values after softmax and GELU functions are quite different from the Gaussian distribution.
We also observe that common quantization metrics, such as MSE and cosine distance, are inaccurate to determine the optimal scaling factor.
In this paper, we propose the twin uniform quantization method to reduce the quantization error on these activation values.
And we propose to use a Hessian guided metric to evaluate different scaling factors, which improves the accuracy of calibration at a small cost.
To enable the fast quantization of vision transformers, we develop an efficient framework, PTQ4ViT.
Experiments show the quantized vision transformers achieve near-lossless prediction accuracy (less than 0.5\% drop at 8-bit quantization) on the ImageNet classification task.
\end{abstract}

%% file: 1_introduction.tex
\section{Introduction}
The self-attention module is the basic building block of the transformer to capture global information~\cite{attention_is_all_you_need_nips2017}.
Inspired by the success of transformers~\cite{BERT_naacl2019,GPT3_nips2020} on natural language processing (NLP) tasks, researchers have brought the self-attention module into computer vision~\cite{ViT_iclr2021,Swin_transformer_iccv2021}.
They replaced the convolution layers in convolutional neural networks (CNNs) with self-attention modules and they called these networks vision transformers.
Vision transformers are comparable to CNNs on many computer vision tasks and have great potential to be deployed on various applications~\cite{Survey_vision_trainsformer_arxiv2021}.

However, both the CNN and the vision transformer are computationally intensive and consume much energy.
The larger and larger scales of neural networks block their deployment on various hardware devices, such as mobile phones and IoT devices, and increase carbon emissions.
It is required to compress these neural networks.
Quantization is one of the most effective ways to compress neural networks~\cite{Survey_quantization_arxiv2021}.
The floating-point values are quantized to integers with a low bit-width, reducing the memory consumption and the computation cost.

There are two types of quantization methods, quantization-aware training (QAT)~\cite{PACT_arxiv2018,LQNets_eccv2018} and post-training quantization (PTQ)~\cite{PTQ_4bit_rapid_deployment_nips2019,Low_bit_quant_iccvw2019}.
Although QAT can generate the quantized network with a lower accuracy drop, the training of the network requires a training dataset, a long optimization time, and the tuning of hyper-parameters.
Therefore, QAT is impractical when the training dataset is not available or rapid deployment is required.
While PTQ quantizes the network with unlabeled calibration images after training, which enables fast quantization and deployment.

\begin{figure}[tbp]
    \centering
    \includegraphics[width=0.55\textwidth]{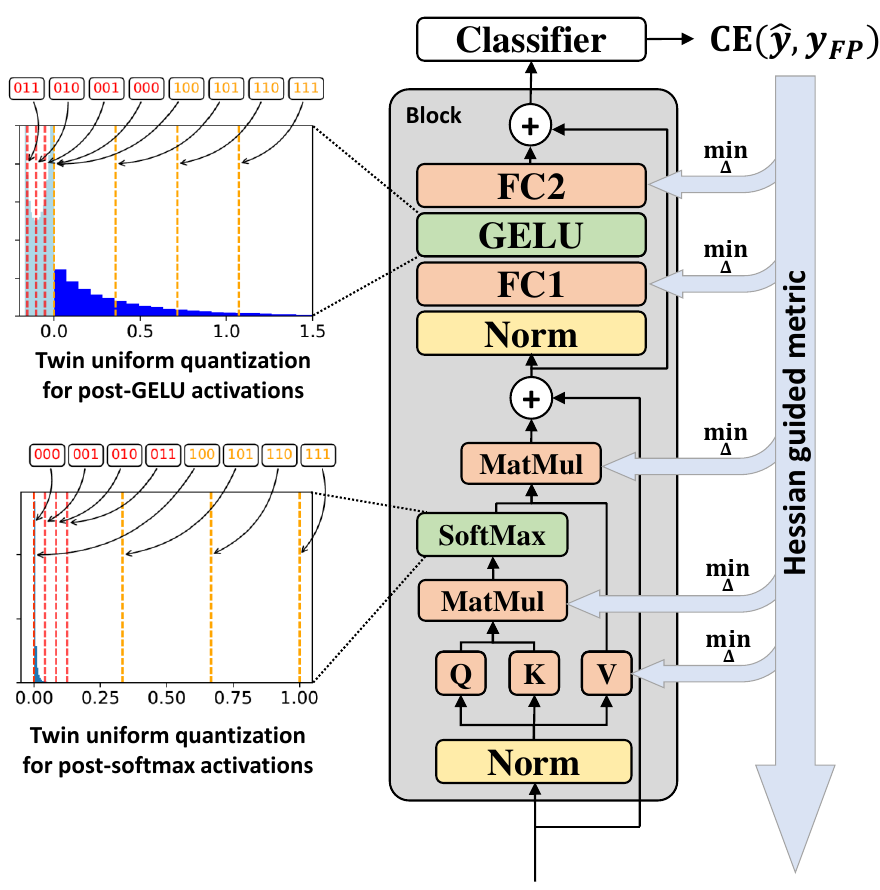}
    \caption{Overview of the PTQ4ViT.}
    \label{overview}
\end{figure} 

Although PTQ has achieved great success on CNNs, directly bringing it to vision transformer results in more than 1\% accuracy drop even with 8-bit quantization~\cite{PTQ_on_ViT_arxiv2021}.
Therefore, we analyze the problems of quantization on vision transformers.
We collect the distribution of activation values in the vision transformer and observe there are some special distributions.
1) The values after softmax have a very unbalanced distribution in $[0,1]$, where most of them are very close to zero.
Although the number of large values is very small, they mean high attention between two patches, which is of vital importance in the attention mechanism.
This requires a large scaling factor to make the quantization range cover the large value.
However, a big scaling factor quantizes the small values to zero, resulting in a large quantization error.
2) The values after the GELU function have an asymmetrical distribution, where the positive values have a large distribution range while the negative values have a very small distribution range.
It's difficult to well quantify both the positive values and negative values with uniform quantization.
Therefore, we propose the twin uniform quantization, which separately quantifies the values in two ranges.
To enable its efficient processing on hardware devices, we design a data format and constrain the scaling factors of the two ranges.

The second problem is that the metric to determine the optimal scaling factor is not accurate on vision transformers.
There are various metrics in previous PTQ methods, including MSE, cosine distance, and Pearson correlation coefficient between the layer outputs before and after quantization.
However, we observe they are inaccurate to evaluate different scaling factor candidates because only the local information is used.
Therefore, we propose to use the Hessian guided metric to determine the quantization parameters, which is more accurate.
The proposed methods are demonstrated in \cref{overview}.

We develop a post-training quantization framework for vision transformers using twin uniform quantization, PTQ4ViT.
\footnote{Code is in \url{https://github.com/hahnyuan/PTQ4ViT}.}
Experiments show the quantized vision transformers (ViT, DeiT, and Swin) achieve near-lossless prediction accuracy (less than 0.5\% drop at 8-bit quantization) on the ImageNet classification task.

Our contributions are listed as follows:
\begin{itemize}
    \item We find the problems in PTQ on vision transformers are special distributions of post-softmax and post-GELU activations and the inaccurate metric.
    \item We propose the twin uniform quantization to handle the special distributions, which can be efficiently processed on existing hardware devices including CPU and GPU.
    \item We propose to use the Hessian guided metric to determine the optimal scaling factors, which replaces the inaccurate metrics.
    \item The quantized networks achieve near-lossless prediction accuracy, making PTQ acceptable on vision transformers.
\end{itemize}

%% file: 2_background_relatedwork.tex
\section{Background and Related Work}

\subsection{Vision Transformer}
In the last few years, convolution neural networks (CNNs) have achieved great success in computer vision.
The convolution layer is a fundamental component of CNNs to extract features using local information.
Recently, the position of CNNs in computer vision is challenged by vision transformers, which take the self-attention modules~\cite{attention_is_all_you_need_nips2017} to make use of the global information. 
DETR~\cite{DETR_eccv2020} is the first work to replace the object detection head with a transformer, which directly regresses the bounding boxes and achieves comparable results with the CNN-based head.
ViT~\cite{ViT_iclr2021} is the first architecture that replaces all convolution layers, which achieves better results on image classification tasks.
Following ViT, various vision transformer architectures have been proposed to boost performance ~\cite{PVT_arxiv2021,Swin_transformer_iccv2021}.
Vision transformers have been successfully applied to downstream tasks~\cite{Swin_transformer_iccv2021,Max_deeplab_cvpr2021,hand_transformer_eccv2020}.
They have great potential for computer vision tasks~\cite{Survey_vision_trainsformer_arxiv2021}.

The input of a transformer is a sequence of vectors. 
An image is divided into several patches and a linear projection layer is used to project each patch to a vector.
These vectors form the input sequence of the vision transformer.
We denote these vectors as $X\in R^{N\times D}$, where $N$ is the number of patches and $D$ is the hidden size, which is the size of the vector after linear projection.

A vision transformer contains some blocks.
As shown in~\cref{overview}, each block is composed of a multi-head self-attention module (MSA) and a multi-layer perceptron (MLP).
MSA generates the attention between different patches to extract features with global information.
Typical MLP contains two fully-connected layers (FC) and the GELU activation function is used after the first layer.
The input sequence is first fed into each self-attention head of MSA.
In each head, the sequence is linearly projected to three matrices, query $Q=XW^Q$, key $K=XW^K$, and value $V=XW^V$.
Then, matrix multiplication $QK^T$ calculates the attention scores between patches.
The softmax function is used to normalize these scores to attention probability $P$.
The output of the head is matrix multiplication $PV$.
The process is formulated as \cref{self-attention}:
\begin{equation}
    \text{Attention}(Q,K,V)=\text{softmax}(\frac{QK^T}{\sqrt{d}})V,
    \label{self-attention}
\end{equation}
where $d$ is the hidden size of head.
The outputs of multiple heads are concatenated together as the output of MSA. 

Vision transformers have a large amount of memory, computation, and energy consumption, which hinders their deployment in real-world applications.
Researchers have proposed a lot of methods to compress vision transformers, such as patch pruning~\cite{Patch_slimming_arxiv2021}, knowledge distillation~\cite{Efficient_vision_transformers_distillation_arxiv2021} ,and quantization~\cite{PTQ_on_ViT_arxiv2021}.

\subsection{Quantization}

Network quantization is one of the most effective methods to compress neural networks.
The weight values and activation values are transformed from floating-point to integer with lower bit-width, which significantly decreases the memory consumption, data movement, and energy consumption.
The uniform symmetric quantization is the most widely used method, which projects a floating-point value $x$ to a $k$-bit integer value $x_q$ with a scaling factor $\Delta$:
\begin{equation}
    x_q=\Psi_k(x,\Delta)=\text{clamp}(\text{round}(\frac{x}{\Delta}),-2^{k-1},2^{k-1}-1),
    \label{sym_quant}
\end{equation}
where $\text{round}$ projects a value to an integer and $\text{clamp}$ constrains the output in the range that $k$-bit integer can represent.
We propose the twin uniform quantization, which separately quantifies the values in two ranges.
\cite{piecewise_linear_quantization_eccv2020} also uses multiple quantization ranges.
However, their method targets CNN and is not suitable for ViT.
They use an extra bit to represent which range is used, taking 12.5\% more storage than our method.
Moreover, they use FP32 computation to align the two ranges, which is not efficient.
Our method uses the shift operation, avoiding the format transformation and extra FP32 multiplication and FP32 addition.

There are two types of quantization methods, quantization-aware training (QAT)~\cite{PACT_arxiv2018,LQNets_eccv2018} and post-training quantization (PTQ)~\cite{PTQ_4bit_rapid_deployment_nips2019,Low_bit_quant_iccvw2019}.
QAT methods combine quantization with network training.
It optimizes the quantization parameters to minimize the task loss on a labeled training dataset.
QAT can be used to quantize transformers~\cite{fully_quantized_transformer_for_machine_transflation_emnlp2020}.
Q-BERT~\cite{Q-BERT_hessian_based_ultra_low_precision_aaai2020} uses the Hessian spectrum to evaluate the sensitivity of the different tensors for mixed-precision, achieving 3-bit weight and 8-bit activation quantization.
Although QAT achieves lower bit-width, it requires a training dataset, a long quantization time, and hyper-parameter tuning.
PTQ methods quantize networks with a small number of unlabeled images, which is significantly faster than QAT and doesn't require any labeled dataset.
PTQ methods should determine the scaling factors $\Delta$ of activations and weights for each layer.
Choukroun et al.~\cite{Low_bit_quant_iccvw2019} proposed to minimize the mean square error (MSE) between the tensors before and after quantization.
EasyQuant~\cite{EasyQuant_arxiv2020} uses the cosine distance to improve the quantization performance on CNN.
Recently, Liu et al.~\cite{PTQ_on_ViT_arxiv2021} first proposed a PTQ method to quantize the vision transformer.
Pearson correlation coefficient and ranking loss are used as the metrics to determine the scaling factors.
However, these metrics are inaccurate to evaluate different scaling factor candidates because only the local information is used.

%% file: 3_method.tex
\section{Method}
In this section, we will first introduce a base PTQ method for vision transformers.
Then, we will analyze the problems of quantization using the base PTQ and propose methods to address the problems.
Finally, we will introduce our post-training quantization framework, PTQ4ViT.

\subsection{Base PTQ for Vision Transformer}
\label{base_ptq}

Matrix multiplication is used in the fully-connected layer and the computation of $QK^T$ and $PV$, which is the main operation in vision transformers.
In this paper, we formulate it as $O=AB$ and we will focus on its quantization.
$A$ and $B$ are quantized to $k$-bit using the symmetric uniform quantization with scaling factors $\Delta_A$ and $\Delta_B$.
According to \cref{sym_quant}, we have $A_q=\Psi_k(A,\Delta_A)$ and $B_q=\Psi_k(B,\Delta_B)$.
In base PTQ, the distance of the output before and after quantization is used as metric to determine the scaling factors, which is formulated as:
\begin{equation}
    \min_{\Delta_A,\Delta_B}\text{distance}(O,\hat{O}),
    \label{optimization}
\end{equation}
where $\hat{O}$ is the output of the matrix multiplication after quantization $\hat{O}=\Delta_A \Delta_B A_q B_q$.

The same as~\cite{EasyQuant_arxiv2020}, we use cosine distance as the metric to calculate the distance.
We make the search spaces of ${\Delta_A}$ and ${\Delta_B}$ by linearly dividing $[\alpha \frac{A_{max}}{2^{k-1}}, \beta \frac{A_{max}}{2^{k-1}}]$ and $[\alpha \frac{B_{max}}{2^{k-1}}, \beta \frac{B_{max}}{2^{k-1}}]$ to $n$ candidates, respectively.
$A_{max}$ and $B_{max}$ are the maximum absolute value of $A$ and $B$.
$\alpha$ and $\beta$ are two parameters to control the search range. 
We alternatively search for the optimal scaling factors ${\Delta_A^*}$ and ${\Delta_B^*}$ in the search space.
Firstly, $\Delta_{B}$ is fixed, and we search for the optimal $\Delta_{A}$ to minimize $\text{distance}(O,\hat{O})$.
Secondly, $\Delta_{A}$ is fixed, and we search for the optimal $\Delta_{B}$ to minimize $\text{distance}(O,\hat{O})$.
$\Delta_{A}$ and $\Delta_{B}$ are alternately optimized for several rounds.

The values of $A$ and $B$ are collected using unlabeled calibration images.
We search for the optimal scaling factors of activation or weight layer-by-layer. 
However, the base PTQ results in more than 1\% accuracy drop on quantized vision transformer in our experiments.


\subsection{Twin Uniform Quantization}


\begin{figure}[t]
  \begin{minipage}[t]{0.5\linewidth} 
    \centering 
    \includegraphics[width=0.9\textwidth]{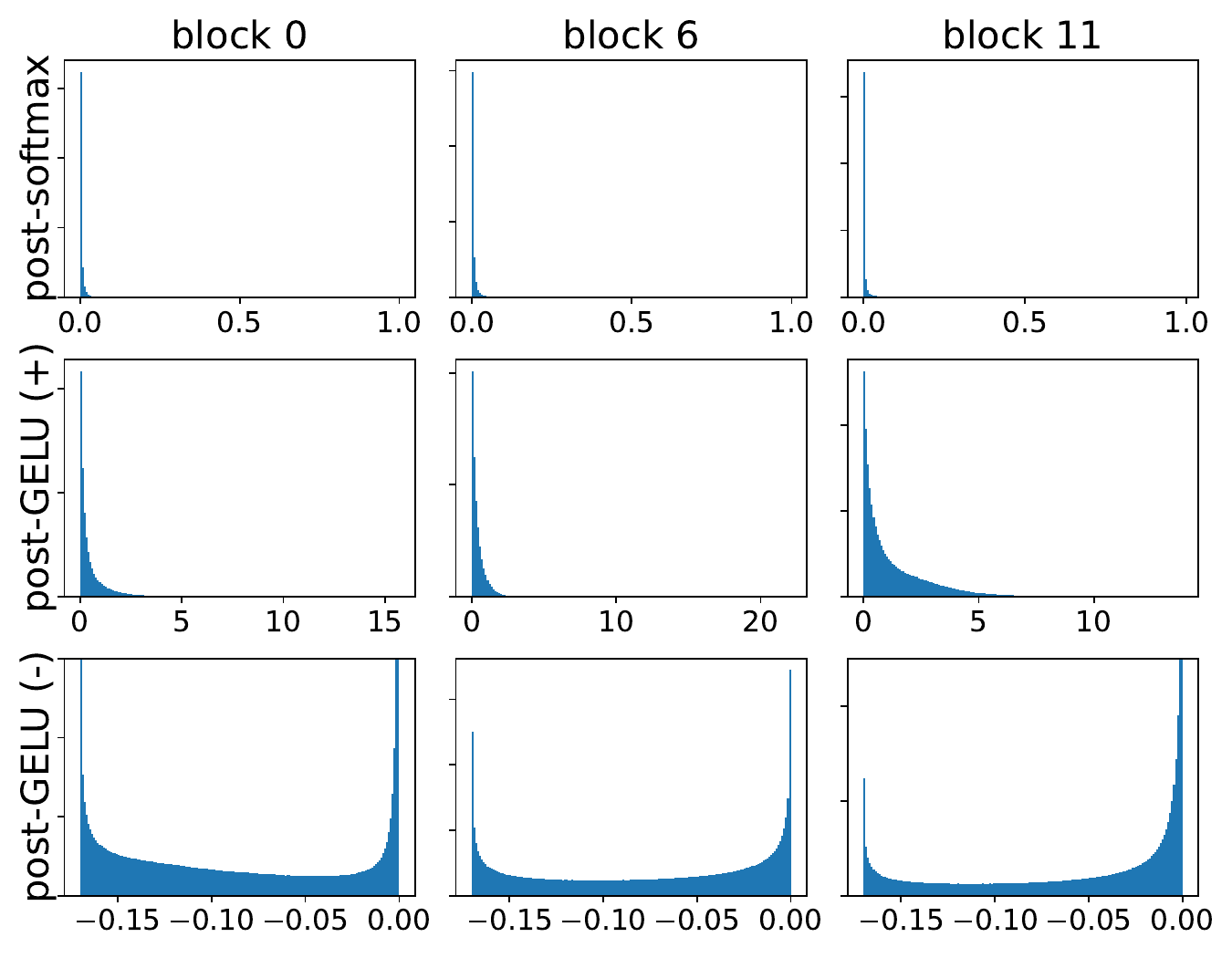}
    \caption{Distributions of the post-softmax values, the positive post-GELU values, and the negative post-GELU values.}
    \label{activation_distribution}
  \end{minipage}
  \begin{minipage}[t]{0.5\linewidth} 
    \centering 
    \includegraphics[width=0.95\textwidth]{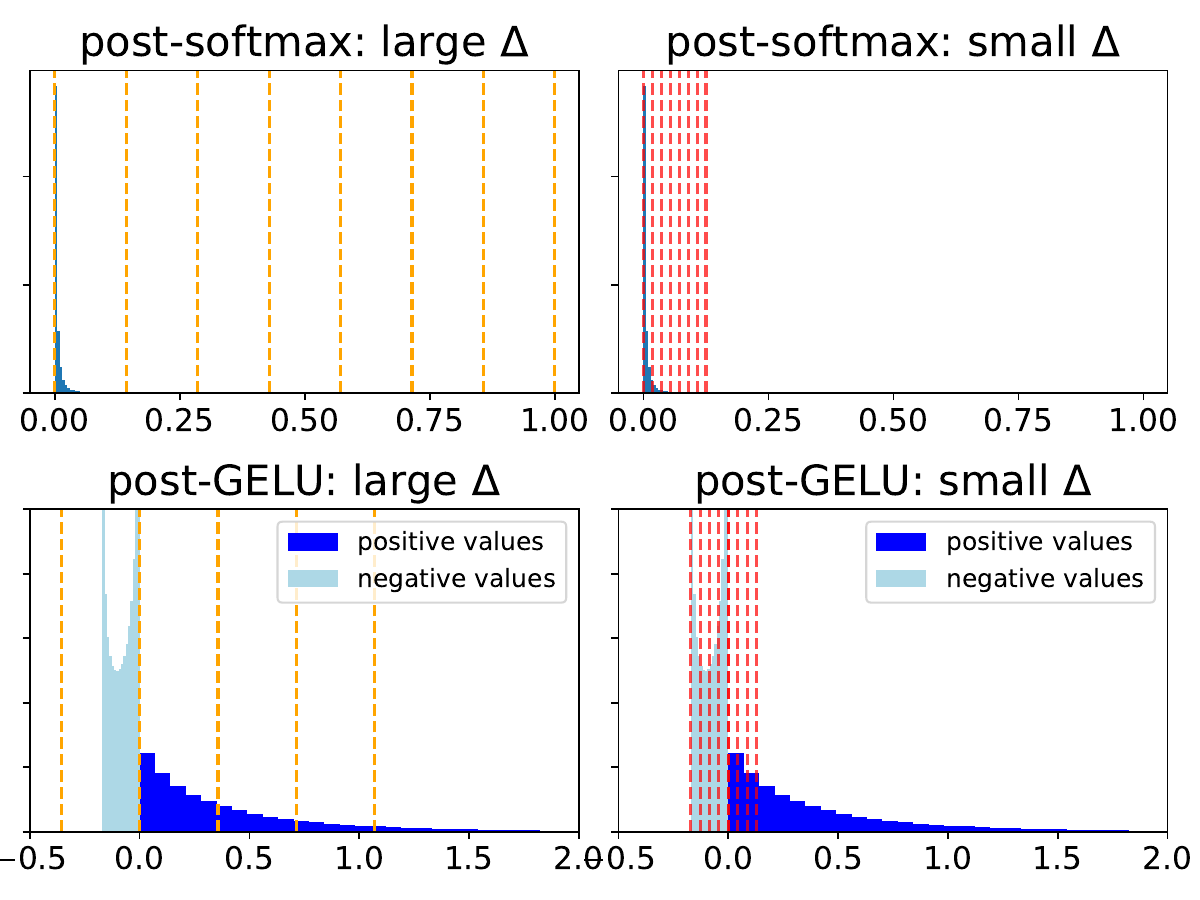}
    \caption{Demonstration of different scaling factors to quantize the post-softmax and post-GELU activation values. 
    }
    \label{scaling_factors}
  \end{minipage} 
\end{figure}

The activation values in CNNs are usually considered Gaussian distributed.
Therefore, most PTQ quantization methods are based on this assumption to determine the scaling factor.
However, we observe the distributions of post-softmax values and post-GELU values are quite special as shown in \cref{activation_distribution}.
Specifically, (1) The distribution of activations after softmax is very unbalanced, in which most values are very close to zero and only a few values are close to one.
(2) The values after the GELU function have a highly asymmetric distribution, in which the unbounded positive values are large while the negative values have a very small distribution range.
As shown in \cref{scaling_factors}, we demonstrate the quantization points of the uniform quantization using different scaling factors.

\begin{figure}[tbp]
    \centering
    \includegraphics[width=0.6\textwidth]{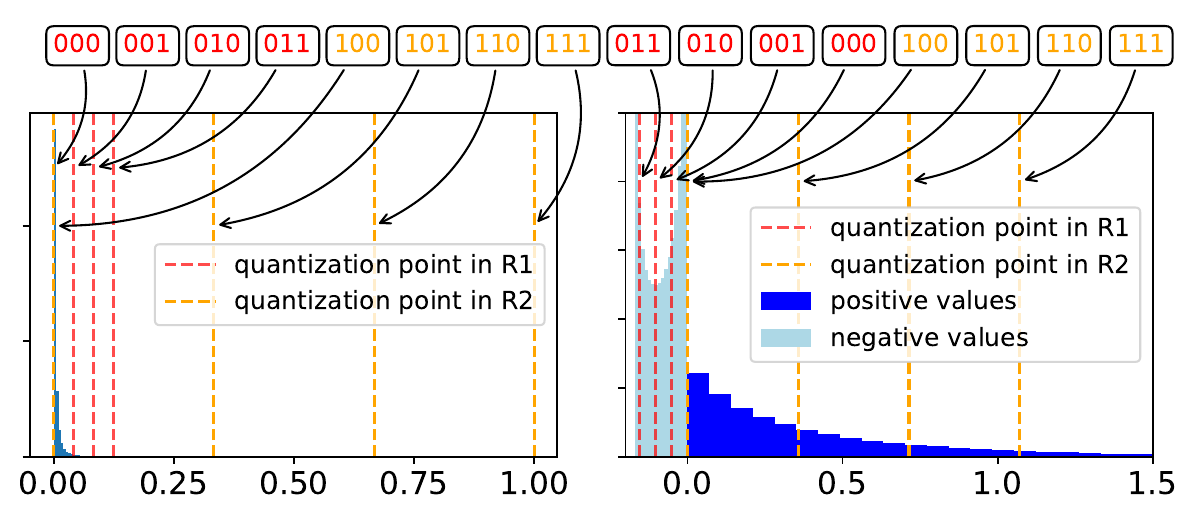}
    \caption{Demonstration of the 3-bit twin uniform quantization on post-softmax values (left) and post-GELU values (right). We annotate the binary values for different quantization points.}
    \label{twin_quant}
\end{figure} 

For the values after softmax, a large value means that there is a high correlation between the two patches, which is important in the self-attention mechanism.
A larger scaling factor can reduce the quantization error of these large values, which causes smaller values to be quantized to zero.
While a small scaling factor makes the large values quantized to small values, which significantly decreases the intensity of attention between two patches.
For the values after GELU, it is difficult to quantify both positive and negative values well with symmetric uniform quantization.
Non-uniform quantization~\cite{network_sketching_cvpr2017} can be used to solve the problem.
It can set the quantization points according to the distribution, ensuring the overall quantization error is small.
However, most hardware devices cannot efficiently process the non-uniform quantized values.
Acceleration can be achieved only on specially designed hardware.

We propose the twin uniform quantization, which can be efficiently processed on existing hardware devices including CPUs and GPUs.
As shown in \cref{twin_quant}, twin uniform quantization has two quantization ranges, R1 and R2, which are controlled by two scaling factors $\Delta_{\text{R1}}$ and $\Delta_{\text{R2}}$, respectively.
The $k$-bit twin uniform quantization is be formulated as:
\begin{equation}
    \text{T}_k(x,\Delta_{\text{R1}},\Delta_{\text{R2}})=\begin{cases}\Psi_{k-1}(x,\Delta_{\text{R1}}), x\in\text{R1} \\ \Psi_{k-1}(x,\Delta_{\text{R2}}), otherwise\end{cases}.
    \label{twin_quant_func}
\end{equation}

For values after softmax, the values in R1 = $[0,2^{k-1}\Delta_{\text{R1}}^{s})$ can be well quantified by using a small $\Delta_{\text{R1}}^{s}$.
To avoid the effect of calibration dataset, we keeps $\Delta_{\text{R2}}^{s}$ fixed to $1/2^{k-1}$.
Therefore, R2 = $[0,1]$ can cover the whole range, and large values can be well quantified in R2. 
For activation values after GELU, negative values are located in R1 = $[-2^{k-1}\Delta_{\text{R1}}^{g},0]$ and positive values are located in R2=$[0, 2^{k-1}\Delta_{\text{R2}}^{g}]$.
We also keep $\Delta_{\text{R1}}^{g}$ fixed to make R1 just cover the entire range of negative numbers.
Since different quantization parameters are used for positive and negative values respectively, the quantization error can be effectively reduced.
When calibrating the network, we search for the optimal $\Delta_{\text{R1}}^{s}$ and $\Delta_{\text{R2}}^{g}$.

The uniform symmetric quantization uses the $k$ bit signed integer data format.
It consists of one sign bit and $k-1$ bits representing the quantity.
In order to efficiently store the twin-uniform-quantized values, we design a new data format.
The most significant bit is the range flag to represent which range is used (0 for R1, 1 for R2).
The other $k-1$ bits compose an unsigned number to represent the quantity.
Because the sign of values in the same range is the same, the sign bit is removed.

Data in different ranges need to be multiplied and accumulated in matrix multiplication.
In order to efficiently process with the twin-uniform-quantized values on CPUs or GPUs, we constrain the two ranges with $\Delta_{\text{R2}}=2^m\Delta_{\text{R1}}$, where $m$ is an unsigned integer.
Assuming $a_q$ is quantized in R1 and $b_q$ is quantized in R2, the two values can be aligned:
\begin{equation}
    a_q\times \Delta_{\text{R1}}+b_q\times \Delta_{\text{R2}}=(a_q+b_q\times 2^{m})\Delta_{\text{R1}}.
\end{equation}
We left shift $b_q$ by $m$ bits, which is the same as multiplying the value by $2^{m}$.
The shift operation is very efficient on CPUs or GPUs.
Without this constraint, multiplication is required to align the scaling factor, which is much more expensive than shift operations.

\subsection{Hessian Guided Metric}
Next, we will analyze the metrics to determine the scaling factors of each layer.
Previous works~\cite{Low_bit_quant_iccvw2019,EasyQuant_arxiv2020,PTQ_on_ViT_arxiv2021} greedily determine the scaling factors of inputs and weights layer by layer.
They use various kinds of metrics, such as MSE and cosine distance, to measure the distance between the original and the quantized outputs.
The change in the internal output is considered positively correlated with the task loss, so it is used to calculate the distance. 

\begin{figure}[tbp]
    \centering
    \includegraphics[width=0.6\textwidth]{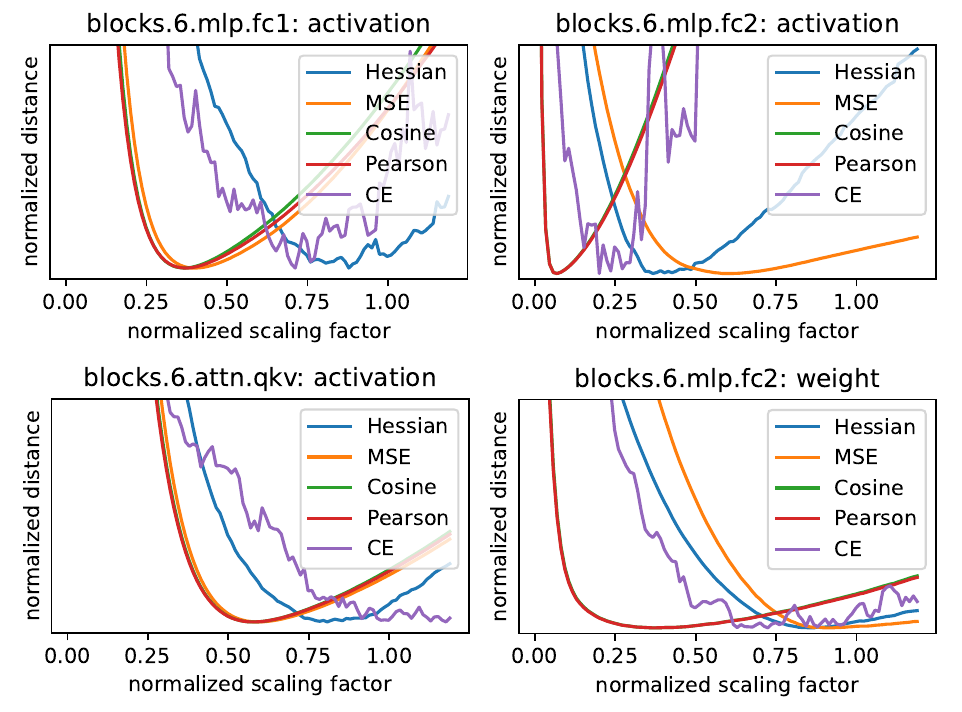}
    \caption{The distance between the layer outputs before and after quantization and the change of task loss (CE) under different scaling factors on ViT-S/224. 
    The x-axis is the normalized scaling factor by dividing $\frac{A_{max}}{2^{k-1}}$ or $\frac{B_{max}}{2^{k-1}}$.}
    \label{inaccurate_metric}
\end{figure} 

We plot the performance of different metrics in \cref{inaccurate_metric}.
We observe that MSE, cosine distance, and Pearson correlation coefficient are inaccurate compared with task loss (cross-entropy) on vision transformers.
The optimal scaling factors based on them are not consistent with that based on task loss.
For instance, on blocks.6.mlp.fc1:activation, they indicate that a scaling factor around $0.4 \frac{A_{max}}{2^{k-1}}$ is the optimal one, while the scaling factor around $0.75 \frac{A_{max}}{2^{k-1}}$ is the optimal according to the task loss.
Using these metrics, we get sub-optimal scaling factors, causing the accuracy degradation.
The distance between the last layer's output before and after quantization can be more accurate in PTQ.
However, using it to determine the scaling factors of internal layers is impractical because it requires executing the network many times to calculate the last layer's output, which consumes too much time.

To achieve high accuracy and quick quantization at the same time, we propose to use the Hessian guided metric to determine the scaling factors.
In the classification task, the task loss is $L=\text{CE}(\hat{y},y)$, where $\text{CE}$ is cross-entropy, $\hat{y}$ is the output of the network, and $y$ is the ground truth\footnote{The ground truth $y$ is not available in PTQ, so we use the prediction of floating-point network $y_{FP}$ to approximate it.}.
When we treat the weights as variables, the expectation of loss is a function of weight $\mathbb{E}[L(W)]$.
The quantization brings a small perturbation $\epsilon$ on weight $\hat{W}=W+\epsilon$.
We can analyze the influence of quantization on task loss by Taylor series expansion.
\begin{equation}
    \mathbb{E}[L(\hat{W})]-\mathbb{E}[L(W)]\approx \epsilon^T \bar{g}^{(W)}+\frac{1}{2}\epsilon^T \bar{H}^{(W)}\epsilon,
\end{equation}
where $\bar{g}^{(W)}$ is the gradients and $\bar{H}^{(W)}$ is the Hessian matrix.
The target is to find the scaling factors to minimize the influence: $\min_{\Delta}(\mathbb{E}[L(\hat{W})]-\mathbb{E}[L(W)])$.
Based on the layer-wise reconstruction method in~\cite{BRECQ_ICLR2021}, the optimization can be approximated\footnote{The derivation of it is in Appendix.} by:
\begin{equation}
\begin{aligned}
     \min_{\Delta}\mathbb{E}[(\hat{O^l}-O^l)^Tdiag\big((\frac{\partial L}{\partial O^l_1})^2 ,\dots, (\frac{\partial L}{\partial O^l_{|O^l|}})^2\big) (\hat{O^l}-O^l)],
\end{aligned}
\label{hessian_optimization}
\end{equation}
where $O^l$ and $\hat{O^l}$ are the outputs of the $l$-th layer before and after quantization, respectively.
As shown in \cref{inaccurate_metric}, the optimal scaling factor indicated by Hessian guided metric is closer to that indicated by task loss (CE).
Although it still has a gap with the task loss, Hessian guided metric significantly improves the performance.
For instance, on blocks.6.mlp.fc1:activation, the optimal scaling factor indicated by Hessian guided metric has less influence on task loss than other metrics.

\subsection{PTQ4ViT Framework}
\label{PTQ4ViT_framework}

\begin{algorithm}[tb]
\small
  \SetAlgoLined
  \caption{Searches for the optimal scaling factors of each layer.}
  \label{ptq4vit}
  \For{$l$ in $1$ to $L$}{
    forward-propagation $O^l\leftarrow A^lB^l$\;
    
  }
  \For{$l$ in $L$ to $1$}{
    backward-propagation to get $\frac{\partial L}{\partial O^l}$\;
  }
  \For{$l$ in $1$ to $L$}{
    initialize $\Delta_{B^l}^*\leftarrow \frac{B^l_{max}}{2^{k-1}}$\;
    generate search spaces of ${\Delta_{A^l}}$ and ${\Delta_{B^l}}$\;
    \For{$r$ = $1$ to $\#\text{Round}$}{
        search for $\Delta_{A^l}^*$ using \cref{hessian_optimization}\;
        search for $\Delta_{B^l}^*$ using \cref{hessian_optimization}\;
    }
  }
\end{algorithm}

    
    

To achieve fast quantization and deployment, we develop an efficient post-training quantization framework for vision transformers, PTQ4ViT.
Its flow is described in \cref{ptq4vit}.
It supports the twin uniform quantization and Hessian guided metric.
There are two quantization phases.
1) The first phase is to collect the output and the gradient of the output in each layer before quantization.
The outputs of the $l$-th layer $O^l$ are calculated through forward propagation on the calibration dataset.
The gradients $\frac{\partial L}{\partial O^l_1} ,\dots, \frac{\partial L}{\partial O^l_a}$ are calculated through backward propagation.
2) The second phase is to search for the optimal scaling factors layer by layer.
Different scaling factors in the search space are used to quantize the activation values and weight values in the $l$-th layer. 
Then the output of the layer $\hat{O^l}$ is calculated.
We search for the optimal scaling factor $\Delta^*$ that minimizes \cref{hessian_optimization}.

In the first phase, we need to store $O^l$ and $\frac{\partial L}{\partial O^l}$, which consumes a lot of GPU memory.
Therefore, we transfer these data to the main memory when they are generated.
In the second phase, we transfer $O^l$ and $\frac{\partial L}{\partial O^l}$ back to GPU memory and destroy them when the quantization of $l$-th layer is finished.
To make full use of the GPU parallelism, we calculate $\hat{O^l}$ and the influence on loss for different scaling factors in batches.

%% file: 4_experiment.tex
\section{Experiments}

In this section, we first introduce the experimental settings.
Then we will evaluate the proposed methods on different vision transformer architectures.
At last, we will take an ablation study on the proposed methods.

\begin{table}[tb]
\setlength\tabcolsep{6pt}
\centering
\caption{Top-1 Accuracy of Quantized Vision Transformers. The result in the bracket is the accuracy drop from floating-point networks. W6A6 means weights are quantized to 6-bit and activations are quantized to 6-bit. The default patch size is 16x16. ViT-S/224/32 means the input resolution is 224x224 and the patch size is 32x32. }
\label{classification}
\begin{tabular}{@{}cccccc@{}}
\toprule
\multirow{2}{*}{Model} & \multirow{2}{*}{FP32} & \multicolumn{2}{c}{Base PTQ} & \multicolumn{2}{c}{PTQ4ViT} \\
                       &                       & W8A8          & W6A6         & W8A8         & W6A6         \\ \midrule
ViT-S/224/32           & 75.99                 & 73.61(2.38)   & 60.14(15.8)  & 75.58(0.41)  & 71.90(4.08)  \\
ViT-S/224              & 81.39                 & 80.46(0.91)   & 70.24(11.1)  & 81.00(0.38)  & 78.63(2.75)  \\
ViT-B/224              & 84.54                 & 83.89(0.64)   & 75.66(8.87)  & 84.25(0.29)  & 81.65(2.89)  \\
ViT-B/384              & 86.00                 & 85.35(0.64)   & 46.88(39.1)  & 85.82(0.17)  & 83.34(2.65)  \\ \midrule
DeiT-S/224             & 79.80                 & 77.65(2.14)   & 72.26(7.53)  & 79.47(0.32)  & 76.28(3.51)  \\
DeiT-B/224             & 81.80                 & 80.94(0.85)   & 78.78(3.01)  & 81.48(0.31)  & 80.25(1.55)  \\
DeiT-B/384             & 83.11                 & 82.33(0.77)   & 68.44(14.6)  & 82.97(0.13)  & 81.55(1.55)  \\ \midrule
Swin-T/224             & 81.39                 & 80.96(0.42)   & 78.45(2.92)  & 81.24(0.14)  & 80.47(0.91)  \\
Swin-S/224             & 83.23                 & 82.75(0.46)   & 81.74(1.48)  & 83.10(0.12)  & 82.38(0.84)  \\
Swin-B/224             & 85.27                 & 84.79(0.47)   & 83.35(1.91)  & 85.14(0.12)  & 84.01(1.25)  \\
Swin-B/384             & 86.44                 & 86.16(0.26)   & 85.22(1.21)  & 86.39(0.04)  & 85.38(1.04)  \\ \bottomrule
\end{tabular}
\end{table}

\subsection{Experiment Settings}
\label{sec:Experiment Settings}

For post-softmax quantization, the search space of $\Delta_{\text{R1}}^s$ is $[\frac{1}{2^{k}},\frac{1}{2^{k+1}},...,\frac{1}{2^{k+10}}]$.
The search spaces of scaling factors for weight and other activations are the same as that of base PTQ (\cref{base_ptq}). 
We set $\ alpha = 0 $, $\ beta = 1.2 $, and $n = 100 $.
The search round $\#Round$ is set to $3$.
We experiment on the ImageNet classification task~\cite{imagenet_IJCV2015}.
We randomly select 32 images from the training dataset as calibration images.
The ViT models are provided by timm~\cite{rw2019timm}.

We quantize all the weights and inputs for the fully-connect layers including the first projection layer and the last prediction layer.
We also quantize the two input matrices for the matrix multiplications in self-attention modules.
We use different quantization parameters for different self-attention heads.
The scaling factors for $W^Q$, $W^K$, and $W^V$ are different.
The same as~\cite{PTQ_on_ViT_arxiv2021}, we don't quantize softmax and normalization layers in vision transformers.

\subsection{Results on ImageNet Classification Task}

We choose different vision transformer architectures, including ViT~\cite{ViT_iclr2021}, DeiT~\cite{DeiT_icml2021}, and Swin~\cite{Swin_transformer_iccv2021}.
The results are demonstrated in \cref{classification}.
From this table, we observe that base PTQ results in more than 1\% accuracy drop on some vision transformers even at the 8-bit quantization.
PTQ4ViT achieves less than 0.5\% accuracy drop with 8-bit quantization.
For 6-bit quantization, base PTQ results in high accuracy drop (9.8\% on average) while PTQ4ViT achieves a much smaller accuracy drop (2.1\% on average).

We observe that the accuracy drop on Swin is not as significant as ViT and DeiT.
The prediction accuracy drops are less than 0.15\% on the four Swin transformers at 8-bit quantization.
The reason may be that Swin computes the self-attention locally within non-overlapping windows.
It uses a smaller number of patches to calculate the self-attention, reducing the unbalance after post-softmax values.
We also observe that larger vision transformers are less sensitive to quantization.
For instance, the accuracy drops of ViT-S/224/32, ViT-S/224, ViT-B/224, and ViT-B/384 are 0.41, 0.38, 0.29, and 0.17 at 8-bit quantization and 4.08, 2.75, 2.89, and 2.65 at 6-bit quantization, respectively.
The reason may be that the larger networks have more weights and generate more activations, making them more robust to the perturbation caused by quantization.

\begin{table}[tb]
\setlength\tabcolsep{6pt}
\centering
\caption{Results of different PTQ methods. \#ims means the number of calibration images. MP means mixed precision. BC means bias correction.}
\label{classification_compare}
\begin{tabular}{@{}cccccc@{}}
\toprule
Model                & Method     & Bit-width  & \#ims & Size & Top-1  \\ \midrule
\multicolumn{1}{l}{} & EasyQuant~\cite{EasyQuant_arxiv2020}  & W8A8      & 1024  & 22.0 & 76.59  \\
\multicolumn{1}{l}{} & Liu~\cite{PTQ_on_ViT_arxiv2021}        & W8A8      & 1024  & 22.0 & 77.47  \\
                     & Liu~\cite{PTQ_on_ViT_arxiv2021}        & W8A8 (MP) & 1024  & 22.2 & 78.09  \\
DeiT-S/224           & PTQ4ViT    & W8A8      & 32    & 22.0 & \textbf{79.47}  \\ \cmidrule(l){2-6} 
79.80                & EasyQuant~\cite{EasyQuant_arxiv2020}  & W6A6      & 1024  & 16.5 & 73.26  \\
                     & Liu~\cite{PTQ_on_ViT_arxiv2021}        & W6A6      & 1024  & 16.5 & 74.58  \\
                     & Liu~\cite{PTQ_on_ViT_arxiv2021}        & W6A6 (MP) & 1024  & 16.6 & 75.10  \\
                     & PTQ4ViT    & W6A6      & 32    & 16.5 & \textbf{76.28}  \\ \midrule
\multicolumn{1}{l}{} & EasyQuant~\cite{EasyQuant_arxiv2020}  & W8A8      & 1024  & 86.0 & 79.36  \\
\multicolumn{1}{l}{} & Liu~\cite{PTQ_on_ViT_arxiv2021}        & W8A8      & 1024  & 86.0 & 80.48  \\
                     & Liu~\cite{PTQ_on_ViT_arxiv2021}        & W8A8 (MP) & 1024  & 86.8 & 81.29  \\
                     & PTQ4ViT    & W8A8      & 32    & 86.0 & \textbf{81.48}  \\ \cmidrule(l){2-6} 
DeiT-B           & EasyQuant~\cite{EasyQuant_arxiv2020}  & W6A6      & 1024  & 64.5 & 75.86  \\
81.80                & Liu~\cite{PTQ_on_ViT_arxiv2021}        & W6A6      & 1024  & 64.5 & 77.02  \\
                     & Liu~\cite{PTQ_on_ViT_arxiv2021}        & W6A6 (MP) & 1024  & 64.3 & 77.47  \\
                     & PTQ4ViT    & W6A6      & 32    & 64.5 & \textbf{80.25}  \\ \cmidrule(l){2-6} 
                     & Liu~\cite{PTQ_on_ViT_arxiv2021}        & W4A4 (MP) & 1024  & 43.6 & \textbf{75.94}  \\
                     & PTQ4ViT & W4A4 & 32 & 43.0 & 60.91 \\
                     & PTQ4ViT+BC & W4A4      & 32    & 43.0 & 64.39 \\ \bottomrule
\end{tabular}
\end{table}

\cref{classification_compare} demonstrates the results of different PTQ methods.
EasyQuant~\cite{EasyQuant_arxiv2020} is a popular post-training method that alternatively searches for the optimal scaling factors of weight and activation.
However, the accuracy drop is more than 3\% at 8-bit quantization.
Liu et al.~\cite{PTQ_on_ViT_arxiv2021} proposed using the Pearson correlation coefficient and ranking loss are used as the metrics to determine the scaling factors, which increases the Top-1 accuracy.
Since the sensitivity of different layers to quantization is not the same, they also use the mixed-precision technique, achieving good results at 4-bit quantization.
At 8-bit quantization and 6-bit quantization, PTQ4ViT outperforms other methods, achieving more than 1\% improvement in prediction accuracy on average.
At 4-bit quantization, the performance of PTQ4ViT is not good.
Although bias correction~\cite{DFQ_bias_correction_iccv2019} can improve the performance of PTQ4ViT, the result at 4-bit quantization is lower than the mixed-precision of Liu et al.
This indicates that mixed-precision is important for quantization with lower bit-width.

\subsection{Ablation Study}

\begin{table}[tb]
\setlength\tabcolsep{6pt}
\caption{Ablation study of the effect of the proposed twin uniform quantization and Hessian guided metric. We mark a $\checkmark$ if the proposed method is used. }
\centering
\label{ablation}
\begin{tabular}{@{}cccccc@{}}
\toprule
\multirow{2}{*}{Model} & \multirow{2}{*}{\begin{tabular}[c]{@{}c@{}}Hessian\\ Guided\end{tabular}} & \multirow{2}{*}{\begin{tabular}[c]{@{}c@{}}Softmax\\ Twin\end{tabular}} & \multirow{2}{*}{\begin{tabular}[c]{@{}c@{}}GELU\\ Twin\end{tabular}} & \multicolumn{2}{c}{Top-1 Accuracy} \\
                       &                                                                           &                                                                         &                                                                      & W8A8             & W6A6            \\ \midrule
\multicolumn{1}{l}{}   & \multicolumn{1}{l}{}                                                      & \multicolumn{1}{l}{}                                                    &                                                                      & 80.47            & 70.24           \\
\multicolumn{1}{l}{}   & $\checkmark$                                                                   &                                                                         &                                                                      & 80.93            & 77.20           \\
ViT-S/224              & $\checkmark$                                                                   & $\checkmark$                                                                 &                                                                      & 81.11            & 78.57           \\
81.39                  & $\checkmark$                                                                   &                                                                         & $\checkmark$                                                              & 80.84            & 76.93           \\
\multicolumn{1}{l}{}   & \multicolumn{1}{l}{}                                                      & $\checkmark$                                                                 & $\checkmark$                                                              & 79.25            & 74.07           \\
\multicolumn{1}{l}{}   & $\checkmark$                                                                   & $\checkmark$                                                                 & $\checkmark$                                                              & 81.00            & 78.63           \\ \midrule
                       & \multicolumn{1}{l}{}                                                      & \multicolumn{1}{l}{}                                                    &                                                                      & 83.90            & 75.67           \\
\multicolumn{1}{l}{}   & $\checkmark$                                                                   &                                                                         &                                                                      & 83.97            & 79.90           \\
ViT-B/224              & $\checkmark$                                                                   & $\checkmark$                                                                 &                                                                      & 84.07            & 80.76           \\
84.54                  & $\checkmark$                                                                   &                                                                         & $\checkmark$                                                              & 84.10            & 80.82           \\
                       & \multicolumn{1}{l}{}                                                      & $\checkmark$                                                                 & $\checkmark$                                                              & 83.40            & 78.86           \\
                       & $\checkmark$                                                                   & $\checkmark$                                                                 & $\checkmark$                                                              & 84.25            & 81.65           \\ \midrule
                       & \multicolumn{1}{l}{}                                                      & \multicolumn{1}{l}{}                                                    &                                                                      & 85.35            & 46.89           \\
\multicolumn{1}{l}{}   & $\checkmark$                                                                   &                                                                         &                                                                      & 85.42            & 79.99           \\
ViT-B/384              & $\checkmark$                                                                   & $\checkmark$                                                                 &                                                                      & 85.67            & 82.01           \\
86.00                  & $\checkmark$                                                                   &                                                                         & $\checkmark$                                                              & 85.60            & 82.21           \\
                       & \multicolumn{1}{l}{}                                                      & $\checkmark$                                                                 & $\checkmark$                                                              & 84.35            & 80.86           \\
                       & $\checkmark$                                                                   & $\checkmark$                                                                 & $\checkmark$                                                              & 85.89            & 83.19           \\ \bottomrule
\end{tabular}
\end{table}

Next, we take ablation study on the effect of the proposed twin uniform quantization and Hessian guided metric. 
The experimental results are shown in \cref{ablation}. 
As we can see, the proposed methods improve the top-1 accuracy of quantized vision transformers.
Specifically, using the Hessian guided metric alone can slightly improve the accuracy at 8-bit quantization, and it significantly improves the accuracy at 6-bit quantization.
For instance, on ViT-S/224, the accuracy improvement is 0.46\% at 8-bit while it is 6.96\% at 6-bit.
And using them together can further improve the accuracy.

Based on the Hessian guided metric, using the twin uniform quantization on post-softmax activation or post-GELU activation can improve the performance.
We observe that using the twin uniform quantization without the Hessian guided metric significantly decreases the top-1 accuracy.
For instance, the top-1 accuracy on ViT-S/224 achieves 81.00\% with both Hessian guided metric and twin uniform quantization at 8-bit quantization, while it decreases to 79.25\% without Hessian guided metric, which is even lower than basic PTQ with 80.47\% top-1 accuracy.
This is also evidence that the metric considering only the local information is inaccurate. 

%% file: 6_conclusion.tex
\section{Conclusion}
In this paper, we analyzed the problems of post-training quantization for vision transformers.
We observed both the post-softmax activations and the post-GELU activations have special distributions.
We also found that the common quantization metrics are inaccurate to determine the optimal scaling factor.
To solve these problems, we proposed the twin uniform quantization and a Hessian-guided metric.
They can decrease the quantization error and improve the prediction accuracy at a small cost.
To enable the fast quantization of vision transformers, we developed an efficient framework, PTQ4ViT.
The experiments demonstrated that we achieved near-lossless prediction accuracy on the ImageNet classification task, making PTQ acceptable for vision transformers.

%% file: 7_acknowledgement.tex
\subsubsection{Acknowledgements} This work is supported by National Key R\&D Program of China (2020AAA0105200), NSF of China (61832020, 62032001, 92064006), Beijing Academy of Artificial Intelligence (BAAI), and 111 Project (B18001).

%% file: 9_appendix.tex
\section{Appendix}

\subsection{Number of Calibration Images}

\begin{table}[tb]
\centering
\caption{Comparing with the time of quantization and the Top-1 accuracy under different number of calibration images with one Nvidia Tesla V100 GPU. T means the quantization time measured in minutes. }
\label{QuantTime}
\begin{tabular}{@{}ccccccc@{}}
\toprule
\multirow{2}{*}{Model} & \multicolumn{3}{c}{\#ims=32} & \multicolumn{3}{c}{\#ims=128} \\
                      & W8A8      & W6A6     & T     & W8A8      & W6A6      & T     \\ \midrule
ViT-S/224/32           & 75.58     & 71.91    & 2     & 75.54     & 72.29     & 5     \\
ViT-S/224              & 81.00     & 78.63    & 3     & 80.99     & 78.44     & 7     \\
ViT-B/224              & 84.25     & 81.65    & 4     & 84.27     & 81.84     & 13    \\
ViT-B/384              & 85.83     & 83.35    & 12    & 85.81     & 83.84     & 43    \\ \midrule
DeiT-S/224             & 79.47     & 76.28    & 3     & 79.41     & 76.51     & 7     \\
DeiT-B/224             & 81.48     & 80.25    & 4     & 81.54     & 80.30     & 16    \\
DeiT-B/384             & 82.97     & 81.55    & 14    & 83.01     & 81.67     & 52    \\ \midrule
Swin-T/224             & 81.25     & 80.47    & 3     & 81.27     & 80.30     & 9     \\
Swin-S/224             & 83.11     & 82.38    & 8     & 83.15     & 82.38     & 17    \\
Swin-B/224             & 85.15     & 84.01    & 10    & 85.17     & 84.15     & 23    \\
Swin-B/384             & 86.39     & 85.39    & 25    & 86.36     & 85.45     & 69    \\ \bottomrule
\end{tabular}
\end{table}

One of the targets of PTQ4ViT is to quickly quantize a vision transformer.
We have proposed to pre-compute the output and gradient of each layer and compute the influence of scaling factor candidates in batches to reduce the quantization time.
As demonstrated in \cref{QuantTime}, PTQ4ViT can quantize most vision transformers in several minutes using 32 calibration images.
Using $\#ims=128$ significantly increases the quantization time.
We observe the Top-1 accuracy varies slightly, demonstrating PTQ4ViT is not very sensitive to $\#ims$.

\subsection{Base PTQ}
Base PTQ is a simple quantization strategy and serves as a benchmark for our experiments. 
Like PTQ4ViT, we quantize all weights and inputs for fully-connect layers (including the first projection layer and the last prediction layer), as well as all input matrices of matrix multiplication operations. 
For fully-connected layers, we use layerwise scaling factors $\Delta_W$ for weight quantization and $\Delta_X$ for input quantization; while for matrix multiplication operations, we use $\Delta_A$ and $\Delta_B$ for A's quantization and B's quantization respectively. 

To get the best scaling factors, we apply a linear grid search on the search space. 
The same as EasyQuant~\cite{EasyQuant_arxiv2020} and Liu et al.~\cite{PTQ_on_ViT_arxiv2021}, we take hyper-parameters $\alpha=0.5$, $\beta = 1.2$, search round $\#Round = 1$ and use cosine distance as the metric. 
Note that in PTQ4ViT, we change the hyper-parameters to $\alpha=0$, $\beta = 1.2$ and search round $\#Round = 3$, which slightly improves the performance.

It should be noticed that Base PTQ adopts a parallel quantization paradigm, which makes it essentially different from sequential quantization paradigms such as EasyQuant~\cite{EasyQuant_arxiv2020}. 
In sequential quantization, the input data of the current quantizing layer is generated with all previous layers quantizing weights and activations. 
While in parallel quantization, the input data of the current quantizing layer is simply the raw output of the previous layer. 

In practice, we found sequential quantization on vision transformers suffers from significant accuracy degradation on small calibration datasets. 
While parallel quantization shows robustness on small calibration datasets. 
Therefore, we choose parallel quantization for both Base PTQ and PTQ4ViT.

\subsection{Derivation of Hessian guided metric}

Our goal is to introduce as small an increment on task loss $L = CE(\hat{y}, y)$ as possible, in which $\hat{y}$ is the prediction of the quantized model and $y$ is the ground truth. 
In PTQ, we don't have labels of input data $y$, so we make a fair assumption that the prediction of floating-point network $y_{FP}$ is close to the ground truth $y$. 
Therefore, we use the $CE(\hat{y}, y_{FP})$ as a substitution of the task loss $L$, which, for convenience, is still denoted as $L$ in the following.

We adopt a layer-wise parallel quantization paradigm. 
We calibrate the scaling factors of a single layer or a single matrix multiplication at a time.
In the following, we derive the Hessian guided metric for the weight, and similar derivations can be applied to activation. 
When we treat the weights of a layer as variables, the expectation of task loss is a function of weight $\mathbb{E}[L(W)]$.
Since quantization on weight introduces a small perturbation $\epsilon$ on weight $\tilde{W} = W + \epsilon$, we could use Taylor series expansion to analyze the influence of quantization on loss.
\begin{equation}
    \mathbb{E}[L(\hat{W})]-\mathbb{E}[L(W)]\approx \epsilon^T \bar{g}^{(W)}+\frac{1}{2}\epsilon^T \bar{H}^{(W)}\epsilon,
    \label{eq:Taylor Expansion of Loss}
\end{equation}
where $\bar{g}^{(W)}$ is the gradients and $\bar{H}^{(W)}$ is the Hessian matrix. 
Since the pretrained model has converged to a local optimum, the gradients $\bar{g}^{(W)}$ is close to zero, thus the first-order term could be ignored and we only consider the second-order term.

The Hessian matrix could be computed by
\begin{equation}
\begin{aligned}
    & \dfrac{\partial^2 L}{\partial w_i \partial w_j} = \dfrac{\partial}{\partial w_j} (\sum_{k=1}^{m} \dfrac{\partial L}{\partial O_k} \dfrac{\partial O_k}{\partial w_i}) \\
    &= \sum_{k=1}^{m} \dfrac{\partial L}{\partial O_k} \dfrac{\partial^2 O_k}{\partial w_i \partial w_j} + \sum_{k,l=1}^{m} \dfrac{\partial O_k}{\partial w_i} \dfrac{\partial^2 L}{\partial O_k \partial O_l} \dfrac{\partial O_l}{\partial w_j},
    \label{eq:Hessian of Weight}
\end{aligned}
\end{equation}
where $O = W^TX \in \mathit{R}^m$ is the output of the layer. 
Note that $\dfrac{\partial^2 O_k}{\partial w_i \partial w_j} = 0$, the first term of \cref{eq:Hessian of Weight} is zero.
We denote $J_{O}(W) $ as the Jacobian matrix of $O$ w.r.t. weight $W$.
Then we have
\begin{equation}
    \bar{H}^{(W)} = J_{O}^T(W) \bar{H}^{(O)} J_{O}(W).
\end{equation}
Since weight's perturbation $\epsilon$ is relatively small, we have a first-order Taylor expansion that $(\hat{O}-O) \approx J_{O}(W) \epsilon$, where $\hat{O}=(W+\epsilon)^TX$.
The second-order term in \cref{eq:Taylor Expansion of Loss} could be written as
\begin{equation}
\begin{aligned}
    \epsilon^T \bar{H}^{(W)}\epsilon & = (J_{O}(W) \epsilon)^T \bar{H}^{(O)} J_{O}(W) \epsilon \\ & \approx (\hat{O}-O)^T\bar{H}^{(O)} (\hat{O}-O)
\end{aligned}
\end{equation}

Next, we introduce how to compute the Hessian matrix of the layer's output $O$. 
Following Liu et al.\cite{BRECQ_ICLR2021}, we use the Fisher Information Matrix $I$ of $O$ to substitute $\bar{H}^{(O)}$. 
For our probabilistic model $p(Y;\theta)$ where $\theta$ is the model's parameters and $Y$ is the random variable of predicted probability, we have:
\begin{equation}
\begin{aligned}
    \bar{I}(\theta) &= \mathbb{E}[\nabla_\theta\log p(Y;\theta) \nabla_\theta\log p(Y;\theta)^T] \\
    &= -\mathbb{E}[\nabla_\theta^2\log p(Y;\theta)]
\end{aligned}
\end{equation}
Notice that $I$ would equal to the expected Hessian $\bar{H}^{(O)}$ if the model's distribution matches the true data distribution, i.e. $y = y_{FP}$. 
We have assumed $y \approx y_{FP}$ in the above derivation, and therefore it is reasonable to replace $\bar{H}^{(O)}$ with $I$.

Using the original Fisher Information Matrix, however, still requires an unrealistic amount of computation.
So we only consider elements on the diagonal, which is $\text{diag}((\dfrac{\partial L}{\partial O_1})^2, \cdots, (\dfrac{\partial L}{\partial O_m})^2)$. 
This only requires the first-order gradient on output $O$, which introduces a relatively small computation overhead.
Therefore, the Hessian guided metric is:
\begin{equation}
    \mathbb{E}[ (\hat{O}-O)^T \text{diag}((\dfrac{\partial L}{\partial O_1})^2, \cdots, (\dfrac{\partial L}{\partial O_m})^2) (\hat{O}-O)].
\end{equation}
Using this metric, we can search for the optimal scaling factor for weight.
The optimization is formulated as:
\begin{equation}
    \mathop{min}_{\Delta_W} \mathbb{E}[ (\hat{O}-O)^T \text{diag}((\dfrac{\partial L}{\partial O_1})^2, \cdots, (\dfrac{\partial L}{\partial O_m})^2) (\hat{O}-O)].
    \label{W_optimization}
\end{equation}
In PTQ4ViT, we make a search space for scaling factors.
Then we compute the influence on the output of the layer $\hat{O}-O$ for each scaling factor.
The optimal scaling factor can be selected according to \cref{W_optimization}.
We assume that $\frac{\partial L}{\partial O}$ doesn't change when the weight is quantized.
This assumption enables the pre-computation of $\frac{\partial L}{\partial O}$, significantly improving the quantization efficiency.

\subsection{More Ablation Study}

We supply more ablation studies for the hyper-parameters.
It is enough to set the number of quantization intervals $\ge$ 20 (accuracy change $< 0.3\%$).
It is enough to set the upper bound of m $\ge$ 15 (no accuracy change).
The best settings of alpha and beta vary from different layers. 
It is appropriate to set $\alpha=0$ and $\beta=1/2^{k-1}$, which has little impact on search efficiency.
We observe that $\#Round$ has little impact on the prediction accuracy (accuracy change $<$ 0.05\% when $\#Round>1$).

We randomly take 32 calibration images to quantize different models 20 times and we observe the fluctuation is not significant. 
The mean/std of accuracies are: ViT-S/32 $75.55\%/0.055\%$ , ViT-S $80.96\%/0.046\%$, ViT-B $84.12\%/0.068\%$, DeiT-S $79.45\%/0.094\%$ , and Swin-S $83.11\%/0.035\%$.